%% file: main.tex
\documentclass[acmtog,nonacm]{acmart}

\input{sections/_preamble.tex}

\input{sections/_template_args.tex}

\begin{document}

\title{Body of Her: A Preliminary Study on End-to-End Humanoid Agent}

\input{sections/0_authors.tex}

\input{sections/0_abstract.tex}
\input{sections/0_article_info.tex}


\maketitle

\input{sections/1_introduction.tex}
\input{sections/3_system_overview.tex}
\input{sections/4_representation.tex}
\input{sections/5_generation.tex}
\input{sections/6_results.tex}
\input{sections/7_limitation}

\begin{acks}
  We would like to thank Zeyi Zhang and Heyuan Yao for their review of this article and their constructive feedback.
  
  It is precisely due to the increasing maturity and lower barriers of the deep learning community in areas such as multi-modal data collection and filtering, foundational models, GPU cluster cloud services, and distributed training frameworks, that the preliminary exploration of the system described in this paper has become feasible with relatively minimal human and financial resources.
\end{acks}

\bibliographystyle{ACM-Reference-Format}
\bibliography{main}

\end{document}

%% file: sections/_preamble.tex
\usepackage{bm}
\usepackage{bbm}

\usepackage{graphicx}
\usepackage{tabularx}
\usepackage{makecell}
\usepackage{subcaption}
\usepackage{multirow}
\usepackage{amsmath}
\usepackage{mathrsfs}

\newcommand{\vect}[1]{\bm{#1}}

\newcommand{\eqword}[1]{{\text{#1}}}

\usepackage{xspace}

\newcommand{\keep}[1]{}
\newcommand{\old}[1]{}

\DeclareMathOperator*{\argmax}{arg\,max}

\newcommand{\norm}[1]{\lVert #1 \rVert}

\usepackage{url}
\usepackage{hyperref}
\newcommand{\fig}{Figure{}~}



%% file: sections/_template_args.tex
\AtBeginDocument{%
  \providecommand\BibTeX{{%
    \normalfont B\kern-0.5em{\scshape i\kern-0.25em b}\kern-0.8em\TeX}}}

\setcopyright{acmlicensed}
\acmJournal{TOG}
\acmYear{0} 
\acmVolume{0} 
\acmNumber{0} 
\acmArticle{0} 
\acmMonth{0} 
\acmPrice{0}
\acmDOI{0}

\acmSubmissionID{0}



%% file: sections/0_authors.tex
\author{Tenglong Ao}
\email{aubrey.tenglong.ao@gmail.com}
\affiliation{
  \institution{Peking University}
  \city{Beijing}
  \country{China}
}

\renewcommand{\shortauthors}{Ao}

%% file: sections/0_abstract.tex
\begin{abstract}
    Interactive virtual humanoid agent is a crucial interface with the physical world. A relatively complete humanoid agent first needs to have face and body, then possess both verbal and non-verbal (such as eye contact, facial expression, lip motion, gesture, and manipulation) abilities, and finally, it is capable of real-time duplex communication, e.g., the ability to actively interrupt conversations. Most prior systems typically only consider a subset of these elements, leaving a gap from realistic humanoid agent. In this work, we propose a real-time, duplex, interactive end-to-end network capable of modeling realistic agent behaviors, including speech, full-body movements for talking, responding, idling, and manipulation. This system is a multimodal model integrating audio and visual inputs, extended from a pre-trained large language model (LLM). We collect approximately 200,000 hours of audio, around 130,000 hours of video data, and about 20,000 alignment samples to build the model. The final model demonstrates capabilities that are difficult to achieve in previous systems, such as generalized object manipulation. This work performs a preliminary exploration of the end-to-end approach in this field, aiming to inspire further research towards scaling up.
\end{abstract}

%% file: sections/0_article_info.tex

\keywords{interactive humanoid agent, multi-modal model, world simulator}

%% file: sections/1_introduction.tex
\section{Introduction}
\label{sec:introduction}
The task of bringing to life a realistic humanoid agent is complex. It involves simultaneously modeling the agent's speech, eye contact, facial expression, lip motion, gesture, and manipulation. Additionally, the agent needs to be capable of real-time perception of signals from the physical world, such as the emotions of the conversation partner, and respond appropriately. To enhance research efficiency, researchers typically divide these elements into different sub-fields for independent parallel research, with the expectation of eventually combining them into a powerful system. However, it often results in a highly complex system that is difficult to optimize, with interfaces between different sub-modules prone to information loss.

A similar situation occurs in other fields, such as natural language processing and autonomous driving. Recently, some large-scale end-to-end systems \cite{brown2020gpt3,tesla2023fsdv12} show superior performances and lead to a convergence of different sub-tasks or sub-modules in these fields. The unified end-to-end network is sufficiently versatile and can be easily scaled up with large-scale data, enabling the joint optimization of complex sub-functions. In this work, we explore to build a unified end-to-end framework for humanoid agent behavior modeling. Our system is duplex, capable of real-time responses to signals from human interlocutors (such as speech and visual cues) and can actively interrupt based on the context of the conversation.

To enable a single network to simultaneously model the agent's speech, eye contact, facial expression, lip motion, gesture, and manipulation, a straightforward idea is to collect a large and diverse set of audio-visual conversation data for large-scale model training. The representation of visual data can be 2D videos or 3D resources (e.g., 3D skeleton-based motion and 3D mesh-based objects). Previous agent-oriented systems \cite{cassell1994rule-basedagent,kopp2006bmlagent,best2020spaagent,cai2024digitallife,ao2023gesturediffuclip} focus on modeling behaviors in the 3D space. But high-quality 3D data is relatively scarce compared to 2D videos, which hinders the system's scaling in terms of the data dimension. Owing to the significant advancements in video generation technology \cite{brooks2024sora}, the utilization of 2D video as a representational format has become feasible. Unlike different 3D elements, which typically employ distinct representations, pixel-based visual representation inherently integrates all visual behaviors of the agent. Additionally, video data is more abundant and easier to obtain. To build the system, we curate approximately 200,000 hours of audio, 130,000 hours of video, and about 20,000 alignment samples. These are used to extend the pre-trained LLM into an audio-vision multimodal model. This model efficiently and comprehensively captures the diverse behaviors of the agent.

In summary, this work performs a preliminary exploration of the unified end-to-end framework for creating the interactive virtual humanoid agent. We explore relatively large-scale training of generative models on audio-vision data. This model is duplex, capable of modeling both natural verbal and non-verbal conversational behaviors of agents. Meanwhile, it can handle some challenging cases, such as manipulating objects and interacting with the surrounding environment within a limited scope. We hope that this human-centric work serves as an entry point for generalized interactive world simulators, inspiring subsequent research to expand the boundary.

%% file: sections/3_system_overview.tex
\section{System Overview}
\label{sec:system_overview}

\begin{figure*}[t]
    \centering
    \includegraphics[width=\linewidth]{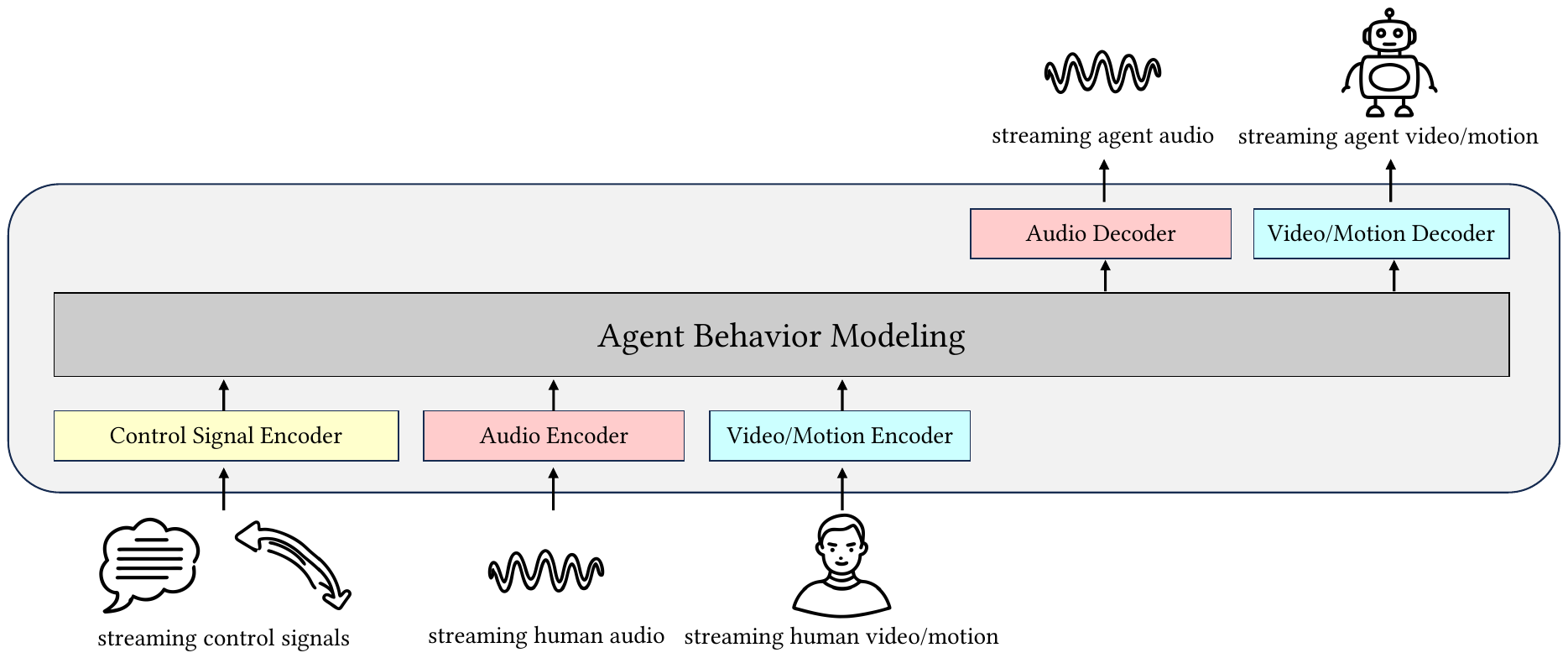}
    \caption{
    Our system continuously synthesizes the agent's voice and visual appearance based on multi-source streaming inputs, including the interlocutor's auditory and visual behaviors and specific control signals. The visual representation can be in video or 3D motion form, depending on rendering and computational power. Control signals use text descriptions for high-level behaviors like emotions and motion trajectories for low-level joint movement guidance.
    }
    \Description{}
    \label{fig:system_overview}
\end{figure*}

Our system aims at modeling the voice and movement behaviors of the agent, which are constantly influenced by the voice, actions, and environment of the human interlocutor. It means that the system is fully duplex. For example, when the human is speaking, the agent needs to simultaneously produce responsive behaviors, and when the agent is speaking, it can simultaneously observe the human's reactions. Additionally, beyond the influence of the interlocutor, our system also supports using motion trajectories and text prompts to affect the agent's behavioral states in real-time.

Specifically, as illustrated in \fig\ref{fig:system_overview}, our system continuously synthesizes the voice and visual appearance of the agent conditioned on multi-source streaming inputs, consisting of the interlocutor's behaviors and some specific control signals. The visual appearance of agent and interlocutor can be represented in the form of video or 3D motion, depending on the rendering method and computational power. To simplify, we default to using video representation in the technical details section. For control signals, we choose text description and motion trajectory to achieve different levels of control. Text prompt can describe high-level behaviors, such as emotions. Motion trajectory specifically refers to the movement trends of body joints, enabling low-level guidance.

%% file: sections/4_representation.tex
\section{Representations of Different Modalities}
\label{sec:representation}

\subsection{Audio Representation}
An efficient and high-fidelity representation of audio is discrete-valued tokens \cite{zeghidour2021soundstream,defossez2022encodec}. A neural network $\mathcal{E}_{A}$ with $K$ residual quantization layers compresses a piece of audio $\vect{A}$ into a discrete-valued token sequence $\vect{Z}^{\eqword{aud}}$ $=$ $[[\vect{z}^{\eqword{aud}}_{k,l}]^{K}_{k=1}]^{L_1}_{l=1}$ as
\begin{align}
    \vect{Z}^{\eqword{aud}} = \mathcal{E}_{A}(\vect{A}),
\end{align}
where each residual quantization layer corresponds to an independent codebook. In this paper, we use the Descript Audio Codec (DAC) \cite{kumar2024dac} as the audio encoder $\mathcal{E}_{A}$, which compresses 44.1kHz audio into discrete-valued tokens at a frame rate of 86Hz and has nine codebooks.

\subsection{Video Representation}
For real-time interaction, it is necessary to compress the video as much as possible while maintaining quality. We have found that utilizing the Querying Transformer \cite{li2023blip2} approach enables the efficient encoding of 2D image features into a compact 1D feature sequence. Recently, TiTok \cite{yu2024titok} explores a similar approach, which can compress a 256 $\times$ 256 $\times$ 3 image into just 32 discrete-valued tokens while retaining competitive performance in visual reconstruction and generation. Our video encoder $\mathcal{E}_{V}$ differs from TiTok in two aspects: (a) to reduce compression losses, we discard the quantizer and switch to continuous representations. And (b) after encoding each frame of the video into continuous-valued tokens, we add a causal temporal transformer \cite{villegas2022phenaki} to model the temporal relationship. This process can be formulated as
\begin{align}
    [\vect{Z}^{\eqword{vid}}_{i}]^{T^{v}}_{i=0} = \mathcal{E}_{V}([\vect{V}_{i}]^{T^{v}}_{i=0}),
\end{align}
where $\vect{V}_{i}$ and $\vect{Z}^{\eqword{vid}}_{i}$ represent the $i$-th video frame and corresponding token sequence, respectively. $\vect{Z}^{\eqword{vid}}_{i}$ consists of $L_{2}$ continuous-valued tokens $[\vect{z}^{\eqword{vid}}_{l}]^{L_{2}}_{l=1}$. $T^{v}$ denotes the video length. Note that we do not compress the temporal dimension of the video. For real-time interaction, the agent's behavior at the $i$-th frame needs to be generated before the arrival of the $i$-th frame from the interlocutor. This requirement is solely related to the original frame rate of the video, so there is no need to increase losses by compressing the temporal dimension. Additionally, compressing $N^{v}$ frames into one would introduce an initial response delay.

Additionally, we can also use 3D human motion to represent the visual behaviors. This representation can be easily integrated into 3D rendering pipeline for the construction of virtual human. A 3D human motion clip is a sequence of poses $\vect{M}$ $=$ $[\vect{m}_{l}]^{L_{2}}_{l=1}$. Each pose $\vect{m}_{l}\in\mathbb{R}^{3+6J}$ is composed of the displacement of the agent and the rotations of its $J$ joints. Since 3D human motion is a simplified representation, we only need to retain the causal temporal transformer in $\mathcal{E}_{V}$ to encode motion into the final latent tokens.

\subsection{Control Signals}
In this paper, we explore two types of control signals: high-level text prompt and low-level motion trajectory. In practice, other control signals can be designed based on specific requirements.

\paragraph{Text Prompt} We use a pre-trained text encoder $\mathcal{E}_{T}$ to encode each text prompt $\vect{P}$. A text prompt typically describes behaviors over a period of time, meaning the prompt remains unchanged for several consecutive frames. Therefore, it is not necessary to modify the prompt frame by frame, allowing us to use a larger text encoder to achieve better performance. We choose the text encoder of a CLIP L/14 model \cite{radford2021clip} as $\mathcal{E}_{T}$, which has been proven effective in the text-to-image domain \cite{esser2024sd3}.

\paragraph{Trajectory} We use the joint velocity to represent the trajectory $\vect{r}$. The velocity of each joint is equal to the difference in corresponding joint positions between two consecutive frames. 
For training data, the joint positions are estimated by an off-the-shelf pose detector \cite{yang2023dwpose}. A linear layer is employed as $\mathcal{E}_{R}$ to encode $\vect{r}$ for the following modeling.

%% file: sections/5_generation.tex
\section{Agent Behavior Modeling}
\label{sec:generation}

\begin{figure*}[t]
    \centering
    \includegraphics[width=\linewidth]{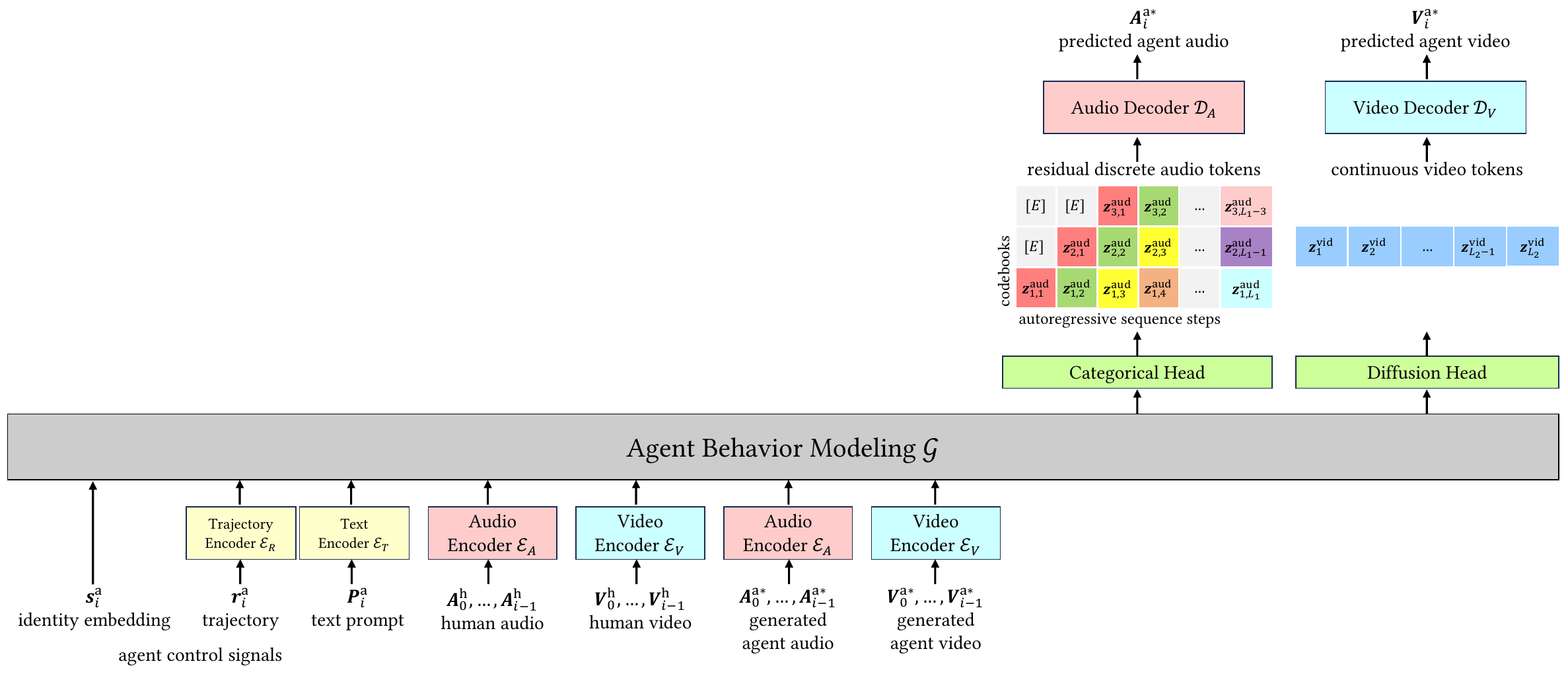}
    \caption{
    A Transformer decoder $\mathcal{G}$ models the probability distribution of agent behaviors at the $i$-th frame, conditioned on previous behaviors of the agent and the human interlocutor, along with specific control signals, as follows: $p_{\mathcal{G}}(\vect{A}^{\eqword{a}}_{i}, \vect{V}^{\eqword{a}}_{i}|\vect{A}^{\eqword{a}}_{<i}, \vect{V}^{\eqword{a}}_{<i}, \vect{C}_{i})$, where $\vect{C}_{i}$ $=$ $[\vect{A}^{\eqword{h}}_{<i}, \vect{V}^{\eqword{h}}_{<i}, \vect{P}^{\eqword{a}}_{i}, \vect{r}^{\eqword{a}}_{i}, \vect{s}^{\eqword{a}}_{i}]$. $\vect{s}^{\eqword{a}}_{i}$ is a learnable embedding that specifies the agent's identity. The predicted agent audio $\vect{A}^{\eqword{a}*}_{i}$ and video $\vect{V}^{\eqword{a}*}_{i}$ are then sampled from this distribution.
    }
    \Description{}
    \label{fig:generator}
\end{figure*}

\subsection{Inference}
As depicted in \fig\ref{fig:generator}, a Transformer decoder $\mathcal{G}$ models the probability distribution of agent behaviors at $i$-th frame conditioned on previous behaviors of the agent and the human interlocutor and specific control signals, $p_{\mathcal{G}}(\vect{A}^{\eqword{a}}_{i}, \vect{V}^{\eqword{a}}_{i}|\vect{A}^{\eqword{a}*}_{<i}, \vect{V}^{\eqword{a}*}_{<i}, \vect{C}_{i})$, where $\vect{C}_{i}$ $=$ $[\vect{A}^{\eqword{h}}_{<i}, \vect{V}^{\eqword{h}}_{<i}, \vect{P}^{\eqword{a}}_{i}, \vect{r}^{\eqword{a}}_{i}, \vect{s}^{\eqword{a}}_{i}]$, $\vect{s}^{\eqword{a}}_{i}$ is a learnable embedding and specifies the agent's identity. Then the predicted agent audio $\vect{A}^{\eqword{a}*}_{i}$ and video $\vect{V}^{\eqword{a}*}_{i}$ are sampled from the distribution.

Specifically, for the predicted agent audio, a standard Softmax-based categorical head is employed to model the probability distribution of discrete audio tokens. We use the delay pattern introduced in \cite{copet2024musicgen} to deal with the hierarchical tokens from different codebooks in the context of an autoregressive model architecture. As for video, we choose the diffusion head proposed by \cite{li2024diffusionhead} to model the 
probability distribution of continuous video tokens. The diffusion head consists of a compact denoising network, such as a MLP. The denoising process models an underlying distribution $p(\vect{z}^{\eqword{vid}}_{l}|\vect{A}^{\eqword{a}*}_{<i}, \vect{V}^{\eqword{a}*}_{<i}, \vect{C}_{i})$ for the output $\vect{z}^{\eqword{vid}}_{l}$. Notably, for each frame, all video tokens $[\vect{z}^{\eqword{vid}}_{l}]^{L_{2}}_{l=1}$ are sampled simultaneously. We can specify a well-designed dialogue scenario as the initial frame $\vect{V}^{\eqword{a}}_{0}$. In practice, we typically do not obtain the initial frame by capturing a real physical scene. A more convenient approach is to directly paste images of the agent and the objects to be manipulated into an empty scene. Then, an image generation model (e.g., Stable Diffusion \cite{rombach2022sd}) is used to refine the composite, ensuring that the pasted elements blend naturally with the empty scene, ultimately resulting in the initial frame.

\subsection{Training}
\label{subsec:training}
Training $\mathcal{G}$ from scratch is expensive. A straightforward idea is to perform modality-adaptation pre-training on a large language model (LLM) without alignment. Since the task scenario falls within the realm of dialogue, it might be possible to skip pre-training and directly perform instruction fine-tuning based on an instruction-tuned LLM for modality extension. A similar idea has been performed in the field of speech language model \cite{zhang2024speechgpt-gen}. The fine-tuning process is divided into two stages: (a) we first fine-tune the LLM with pure audio dialogue data, making it a qualified large speech model (LSM); Then (b) we use synchronized audio-visual dialogue data to fine-tune the LSM, finally extending it to a large speech-video model (LSVM).

$\mathcal{G}$ is first initialized by an instruction-tuned LLM (e.g., Qwen2-7B-Instruct \cite{qwen2}). Then we construct a dataset $\mathcal{C}^{\eqword{aud}}$ $=$ $\{[\vect{A}^{\eqword{a}}_{i,j},$ $\vect{A}^{\eqword{h}}_{i,j},$ $\vect{P}^{\eqword{a}}_{i,j}]^{N_{j}-1}_{i=0}\}^{N}_{j=1}$, which consists of synchronized speech from both parties in the dialogue and textual descriptions of the agent's speech style. Other information not included, such as $\vect{r}$, is set as the empty token. The training objective is to minimize the negative log-likelihood, which can be formulated as 
\begin{align}
    \mathcal{L}(\mathcal{G}|\mathcal{C}^{\eqword{aud}}) = - \sum^{N}_{j=1} \sum^{N_{j}-1}_{i=1} \log p_{\mathcal{G}}(\vect{Z}^{\eqword{aud(a)}}_{i,j}|\vect{A}^{\eqword{a}}_{<i,j}, \vect{A}^{\eqword{h}}_{<i,j}, \vect{P}^{\eqword{a}}_{<i,j}).
\end{align}
Regarding the construction of the dataset, we first collect data from sources such as podcasts, interviews, debates, and open-source dialogue datasets (e.g.,  Gigaspeech \cite{chen2021gigaspeech}). Among these, debate data helps the model learn how to naturally continue a conversation or actively interrupt the other party. Next, we also create some synthetic data: (a) we train a text-to-speech (TTS) model, optimized for dialogue scenarios, to convert text dialogues, such as interview transcripts, into dialogue audio; (b) Most audio, especially from public datasets, is recorded in the form of a single person describing an object or telling a story. We use a language model (e.g., GPT-4o \cite{openai2024gpt4o}) to generate inquiry content based on these audio recordings and then convert it into speech to create dialogue audio data. It is challenging to synthesize data where two people are speaking simultaneously, so we collect more such data from the internet. Finally, we use several speech captioners (e.g., SALMONN \cite{tang2024salmonn}) to annotate all the collected data. The dataset comprises approximately 200,000 hours of audio.

Next, we construct an audio-visual dialogue dataset $\mathcal{C}^{\eqword{vid}}$ $=$ $\{[\vect{A}^{\eqword{a}}_{i,j},$ $\vect{V}^{\eqword{a}}_{i,j},$ $\vect{C}_{i,j}]^{N_{j}-1}_{i=0}\}^{N}_{j=1}$, where $\vect{C}_{i,j}$ $=$ $[\vect{A}^{\eqword{h}}_{<i,j}, \vect{V}^{\eqword{h}}_{<i,j}, \vect{P}^{\eqword{a}}_{i,j}, \vect{r}^{\eqword{a}}_{i,j}, \vect{s}^{\eqword{a}}_{i,j}]$. The training objective can be formulated as the hybrid of a negative log-likelihood and a denoising criterion
\begin{align}
    \mathcal{L}(\mathcal{G}|\mathcal{C}^{\eqword{vid}}) &= - \sum^{N}_{j=1} \sum^{N_{j}-1}_{i=1} \log p_{\mathcal{G}}(\vect{Z}^{\eqword{aud(a)}}_{i,j}|\vect{A}^{\eqword{a}}_{<i,j}, \vect{C}_{<i,j}) \nonumber \\
    &+ \norm{\vect{\epsilon}_{i,j} - \vect{\epsilon}_{\theta}(\vect{Z}^{\eqword{vid(a)}}_{i,j,t_{i,j}}|t_{i,j}, \mathcal{G}^{\eqword{vid}}(\vect{A}^{\eqword{a}}_{<i,j}, \vect{C}_{<i,j}))}^{2},
\end{align}
where $\vect{\epsilon}_{i,j}$ is a noise vector sampled from $\mathcal{N}(\vect{0}, \vect{I})$. $\vect{\epsilon}_{\theta}$ is the diffusion head, parameterized by $\theta$. $t_{i,j}$ is a time step of the noise schedule. $\mathcal{G}^{\eqword{vid}}(\vect{A}^{\eqword{a}}_{<i,j}, \vect{C}_{<i,j})$ means the video features of the output of the generator $\mathcal{G}$ at the $i$-th frame in the sample $j$.

We divide the construction process of $\mathcal{C}^{\eqword{vid}}$ into four stages:
\begin{itemize}
    \item \textbf{Third-person perspective data:} Common sources of dialogue videos include video podcasts, movies, TV shows, and video interviews. The dialogues in these videos are usually filmed from independent third-person perspectives. Although what we need are first-person perspective dialogue videos, this type of third-person perspective data is abundant and can serve as a form of weak supervision. We collect some of this data and use it at the early stage during fine-tuning. The control signals extracted from this data are inaccurate, so we set $\vect{s}$, $\vect{r}$, and $\vect{P}$ to empty tokens during fine-tuning on this data.
    \item \textbf{First-person perspective data:} We collect high-quality first-person dialogue data from scenarios such as online video collaboration, talk show programs, dual-host shows, and specific-perspective dual-person podcasts. This data is used in the later stage of fine-tuning, and control signals $\vect{r}$ and $\vect{P}$ are extracted from it. We employ some vision-language models (VLM) (e.g. GPT-4o \cite{openai2024gpt4o}) to describe the appearance of the interlocutors, their environment, and the actions represented by the selected keyframe, among other details, as $\vect{P}$.
    \item \textbf{Synthetic data:} We construct three types of synthetic data: (a) We collect a large number of single-speaker videos, such as public speeches, news broadcasts, online live streams, etc. We then use a LLM to generate inquiry statements based on the content, forming them into dialogues. The inquiry statements are converted into speech using a TTS model. During training, $\vect{s}$ and $\vect{V}^{\eqword{h}}$ are set as empty tokens. (b) Modeling realistic idle behavior is crucial. We collect some human idle segments and use a text-to-audio model (e.g., AudioGen \cite{kreuk2022audiogen}) to generate ambient sounds. During training, we randomly select idle segments from two different speakers and splice them together into a dialogue video, setting $\vect{s}$ as empty tokens. (c) To enhance the model's performance in manipulation tasks, we reconstruct a virtual agent from datasets (e.g., \cite{banerjee2024hot3d}) containing high-quality 3D annotations of hand manipulation and head movements/gaze signals and other sparse markers. This allow us to obtain $\vect{V}^{\eqword{a}}$ through 3D rendering. Then, we use a VLM to describe the agent's behavior in text, and the LLM generate a spoken script for the agent based on this description, resulting in $\vect{A}^{\eqword{a}}$. As for $\vect{A}^{\eqword{h}}$, it can be obtained similarly to (a). During training, $\vect{s}$ and $\vect{V}^{\eqword{h}}$ are set as empty tokens.
    \item \textbf{Customized agent data:} If we want to "clone" a specific person, we can specifically collect high-quality first-person dialogue data from that person and use it for fine-tuning at the final stage (specifying a speaker embedding $\vect{s}$ during training). To achieve stable results, we find that it is necessary to collect at least several hours of data for each person. A more flexible approach is to create a specific agent through just a picture prompt or a short video prompt. However, it is challenging to achieve satisfactory quality with our limited computational power, and it may require more powerful, larger-scale models to accomplish it.
\end{itemize}
After filtering using the methods referenced in \cite{blattmann2023stablevideodiffusion}, except for customized agent data, the final dataset contains approximately 130,000 hours of video with varying frame rates, resolutions, and aspect ratios. The data of the first three stages account for 15$\%$, 50$\%$, and 35$\%$ of the total, respectively.

\subsection{Reinforcement Learning with Human Feedback (RLHF)}
We employ a RLHF procedure to further align model behavior with human preferences. To collect binary human preference data, we ask annotators to first engage in one question-answer round with an agent. The audio-visual data performed by the annotators is then used as a prompt to sample another segment of the agent's data from a different model variant with varying temperature hyper-parameter. Afterwards, the annotators select the preferred segment from the results. We focus on the agent's performance in five dimensions: speaking, listening, manipulation, identity consistency, and question-answer safety. Therefore, the annotators' interactions and ranking guidance are set up around these dimensions. In total, we collect approximately 20,000 samples of binary comparisons.

We first train a reward model, which takes a sample of audio-visual data from both parties in a conversation as inputs and outputs a scalar score to indicate the quality of the model's generation. We initialize our reward models from the fine-tuned model $\mathcal{G}$ in Section \ref{subsec:training}, ensuring that both models leverage the knowledge gained during the fine-tuning process. The model architecture and hyper-parameters are identical for both models, with the only difference being that the head for next-token prediction is replaced with a regression head to output a scalar reward. Similar to \cite{ouyang2022rlhf}, the training objective of the reward model is formulated as
\begin{align}
    \mathcal{L}_{\eqword{reward}} = - \log(\sigma(r_{\theta}(\vect{Y}_{w}) - r_{\theta}(\vect{Y}_{l}))),
\end{align}
where $r_{\theta}(\vect{Y})$ is the scalar score assigned to a sample of dialogue data $\vect{Y}$ $=$ $[\vect{A}^{\eqword{h}}_{i},$ $\vect{V}^{\eqword{h}}_{i},$ $\vect{A}^{\eqword{a}}_{i},$ $\vect{V}^{\eqword{a}}_{i}]^{N-1}_{i=0}$, with model weight $\theta$. $\vect{Y}_{w}$ represents the sample preferred by annotators, while $\vect{Y}_{l}$ is the rejected sample.

Then, we use proximal policy optimization (PPO) algorithm \cite{schulman2017ppo} for RLHF fine-tuning. We utilize the reward model to approximate the true reward function $\mathcal{R}(\cdot)$ and consider the pre-trained agent model as the policy $\mathcal{\pi}$ that we aim to optimize. The optimization objective is
\begin{align}
    \argmax_{\mathcal{\pi}} \mathbb{E}_{\vect{Y}^{\eqword{h}} \sim \mathcal{C}, \vect{Y}^{\eqword{a}} \sim \mathcal{\pi}}[\mathcal{R}(\vect{Y}^{\eqword{a}}|\vect{Y}^{\eqword{h}})],
\end{align}
where $\vect{Y}^{\eqword{h}}$ and $\vect{Y}^{\eqword{a}}$ represent audio-visual data of the interlocutor and agent, which are sampled from the dataset $\mathcal{C}$ and the policy $\mathcal{\pi}$, respectively. For training stability, following \cite{stiennon2020learninghf,ouyang2022rlhf}, the final reward function $\mathcal{R}(\cdot)$ includes a penalty term for deviating from the original policy $\mathcal{\pi}_{0}$, which is formulated as
\begin{align}
    \mathcal{R}(\vect{Y}^{\eqword{a}}|\vect{Y}^{\eqword{h}}) = \mathcal{R}(\vect{Y}^{\eqword{a}}|\vect{Y}^{\eqword{h}}) - \beta D_{\eqword{KL}}(\mathcal{\pi}_{\theta}(\vect{Y}^{\eqword{a}}|\vect{Y}^{\eqword{h}}) \parallel \mathcal{\pi}_{0}(\vect{Y}^{\eqword{a}}|\vect{Y}^{\eqword{h}})),
\end{align}
where $\beta$ is set as 0.01 in the paper.

%% file: sections/6_results.tex
\section{Results}
\label{sec:results}

\subsection{Details}
Our system outputs video with synchronized audio at a frame rate of 24 in an autoregressive manner. Our model can sample widescreen 640 $\times$ 360 videos, vertical 360 $\times$ 640 videos and everything inbetween. For real-time interaction, the model ($\sim$ 8 billion parameters) is optimized to generate the next frame within 42 milliseconds ($\approx 1 / 24$). For training, inspired by \cite{dehghani2023scalingvit,henry2020qknorm}, we employ bfloat16 and QK norm for stabilize training on a large scale. Regarding GPU usage, taking the NVIDIA A800 80G GPU cluster as an example, the combined duration for both the pre-training and post-training phases is approximately 23 days.

\subsection{Enhancing Reasoning via Chain-of-Thought (CoT)}
\begin{figure*}[t]
    \centering
    \includegraphics[width=\linewidth]{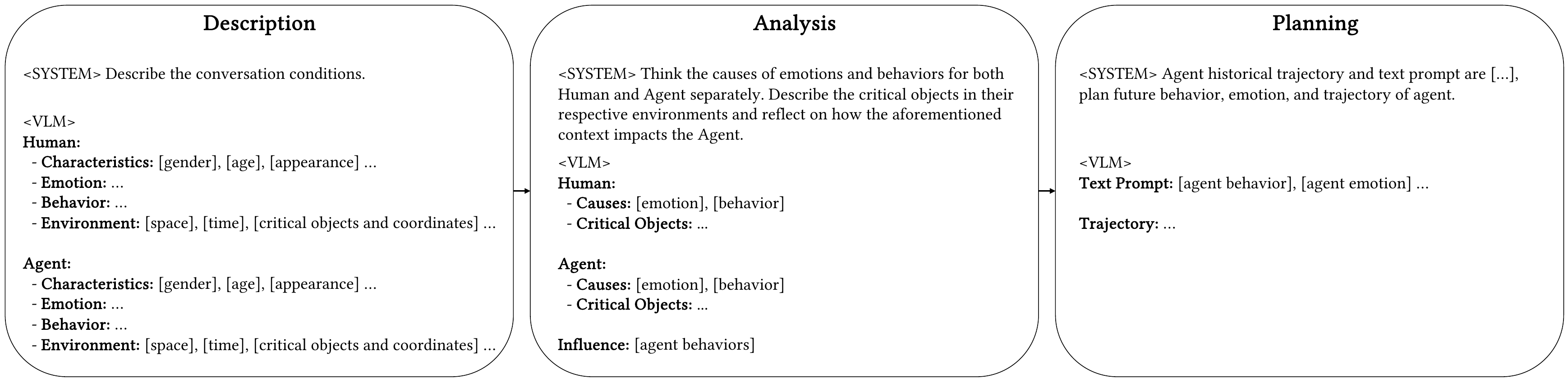}
    \caption{
    Chain-of-thought (CoT) process with the vision-language model (VLM): given historical conversation context, i.e., image key-frames and transcript (transcribed from audio), (a) the \emph{description} stage linguistically portrays the characteristics (e.g., gender, age, and appearance), emotion, behavior, environment (e.g., space, time, critical objects) of the human and the agent, respectively. (b) The \emph{analysis} stage focuses on details of emotions, behaviors, and critical objects and their influence on the agent. (c) The \emph{planning} stage makes final predictions of the future behavior, emotion, and trajectory of agent. The predicted text prompt and trajectory are utilized as control signals of $\mathcal{G}$ to affects agent's future behaviors.
    }
    \Description{}
    \label{fig:system2}
\end{figure*}

Inspired by \cite{tian2024drivevlm}, we can extend current system into a dual system by adding a vision-language model (VLM) for enhanced understanding and planning. The VLM contains a chain-of-thought process with three stages: \emph{Description}, \emph{Analysis}, and \emph{Planning}. Specifically, as shown in \fig\ref{fig:system2}, given historical conversation context, i.e., image key-frames and transcript (transcribed from audio), the description stage linguistically portrays the characteristics (e.g., gender, age, and appearance), emotion, behavior, environment (e.g., space, time, critical objects) of the human and the agent, respectively. The analysis stage focuses on details of emotions, behaviors, and critical objects and their influence on the agent. The planning stage makes final predictions of the future behavior, emotion, and trajectory of agent. The predicted text prompt and trajectory are utilized as control signals of $\mathcal{G}$ to affects agent's future behaviors.

We need to collect data in the format shown in \fig\ref{fig:system2} for fine-tuning the VLM. Due to limited resources, we directly use GPT-4o \cite{openai2024gpt4o} to analyze the annotated data for training $\mathcal{G}$, and automatically construct the formatted fine-tuning data. A potential future improvement could involve manual filtering to enhance data quality. The final dataset contains about 3,000 samples. We use Qwen-VL ($\sim$ 10 billion parameters) \cite{bai2023qwenvl} as the initial VLM, inputting four frames of historical video sampled every second from the current moment, and the model ultimately predicts the text prompt and trajectory for the future 3 seconds. The average inference speed of the fine-tuned VLM is between 500 ms and 1 s, and it runs asynchronously with $\mathcal{G}$.

\subsection{Qualitative Results}
We collect 9 hours and 5.5 hours of curated audio-visual dialogue data from two actors respectively to create corresponding agents. To protect personal privacy, the head part of characters in all demonstrated results are covered with animated Memoji, and the audio is modified to disguise their voices. Notably, during testing, both the topics of conversations with the agent and the objects the agent interacts with are out-of-distribution (OOD), meaning that the test cases have never appeared in the corresponding customized agent dataset. The model may have been exposed to functionally similar scenarios and objects during the pre-training phase, thereby knowing how to interact with them. Some demonstrations of our system can be found \href{https://aubrey-ao.github.io/BodyOfHer}{\textcolor{blue}{here}}\footnote{https://aubrey-ao.github.io/BodyOfHer}.

%% file: sections/7_limitation.tex
\section{Limitations}
\label{sec:limitations}

There are some limitations in our system. First, constrained by model size, there may be issues with logical inconsistencies in the question-and-answer process. The expansion of model size and the requirement for real-time inference are inherently contradictory. As computational power and inference techniques advance, it is expected to be alleviated.

Second, to ensure long-term identity consistency, the appearance of the agent, such as clothing, must align with the corresponding agent's training data. It is challenging to re-edit the agent's attire as needed. In the future, a more flexible approach is to create a specific agent through a single image prompt or a short video prompt. 

Third, the generated results exhibit inconsistencies with the laws of physics. For instance, (a) In \href{https://aubrey-ao.github.io/BodyOfHer}{\textcolor{blue}{Demo1}}, vitamin C should be taken out of the box rather than appearing spontaneously from behind the box; (b) In \href{https://aubrey-ao.github.io/BodyOfHer}{\textcolor{blue}{Demo10}}, after the second cutting, the paper strip reattaches itself automatically; (c) Sometimes, the agent does not strictly have five fingers on one of its hands.

Fourth, the agent exhibits spatial disorientation regarding left and right. For example, in \href{https://aubrey-ao.github.io/BodyOfHer}{\textcolor{blue}{Demo7}}, the curtain should be positioned to the right of the agent rather than to the left.

Fifth, in the real world, there are numerous interactive objects. But due to limitations in the dataset and computational resources, the interaction process of the agent may exhibit artifacts.

Last, the primary reason for the system's capability to produce highly consistent video results is the limited scope of the modeled scenarios, mainly focusing on "two-person dialogues in static scenes." While ensuring real-time performance, we believe that with the expansion of the model and data scale, this paradigm has the potential to address more complex and generalized scenarios, such as intricate dynamic scenes, high-degree-of-freedom agent navigation, multi-agent interactions, and dialogues involving non-humanoid agents.